\documentclass[letterpaper, 10 pt, conference]{ieeeconf}  

\IEEEoverridecommandlockouts                              

\usepackage{tex-utils/helper}
\usepackage{tex-utils/helper_math}
\usepackage{tex-utils/helper_glossaries}
\usepackage{units}
\usepackage{booktabs}
\usepackage{amssymb}

\crefname{figure}{Fig.}{Figs.}
\Crefname{figure}{Fig.}{Figs.}
\crefname{equation}{Eq.}{Eqs.}
\Crefname{equation}{Eq.}{Eqs.}
\crefname{section}{Sec.}{Secs.}
\Crefname{section}{Sec.}{Secs.}

\let\vec\bm
\usepackage[normalem]{ulem}

\makeatletter
\let\save@mathaccent\mathaccent
\newcommand*\if@single[3]{%
  \setbox0\hbox{${\mathaccent"0362{#1}}^H$}%
  \setbox2\hbox{${\mathaccent"0362{\kern0pt#1}}^H$}%
  \ifdim\ht0=\ht2 #3\else #2\fi
  }
\newcommand*\rel@kern[1]{\kern#1\dimexpr\macc@kerna}
\newcommand*\widebar[1]{\@ifnextchar^{{\wide@bar{#1}{0}}}{\wide@bar{#1}{1}}}
\newcommand*\wide@bar[2]{\if@single{#1}{\wide@bar@{#1}{#2}{1}}{\wide@bar@{#1}{#2}{2}}}
\newcommand*\wide@bar@[3]{%
  \begingroup
  \def\mathaccent##1##2{%
    \let\mathaccent\save@mathaccent
    \if#32 \let\macc@nucleus\first@char \fi
    \setbox\z@\hbox{$\macc@style{\macc@nucleus}_{}$}%
    \setbox\tw@\hbox{$\macc@style{\macc@nucleus}{}_{}$}%
    \dimen@\wd\tw@
    \advance\dimen@-\wd\z@
    \divide\dimen@ 3
    \@tempdima\wd\tw@
    \advance\@tempdima-\scriptspace
    \divide\@tempdima 10
    \advance\dimen@-\@tempdima
    \ifdim\dimen@>\z@ \dimen@0pt\fi
    \rel@kern{0.6}\kern-\dimen@
    \if#31
      \overline{\rel@kern{-0.6}\kern\dimen@\macc@nucleus\rel@kern{0.4}\kern\dimen@}%
      \advance\dimen@0.4\dimexpr\macc@kerna
      \let\final@kern#2%
      \ifdim\dimen@<\z@ \let\final@kern1\fi
      \if\final@kern1 \kern-\dimen@\fi
    \else
      \overline{\rel@kern{-0.6}\kern\dimen@#1}%
    \fi
  }%
  \macc@depth\@ne
  \let\math@bgroup\@empty \let\math@egroup\macc@set@skewchar
  \mathsurround\z@ \frozen@everymath{\mathgroup\macc@group\relax}%
  \macc@set@skewchar\relax
  \let\mathaccentV\macc@nested@a
  \if#31
    \macc@nested@a\relax111{#1}%
  \else
    \def\gobble@till@marker##1\endmarker{}%
    \futurelet\first@char\gobble@till@marker#1\endmarker
    \ifcat\noexpand\first@char A\else
      \def\first@char{}%
    \fi
    \macc@nested@a\relax111{\first@char}%
  \fi
  \endgroup
}
\makeatother

\usepackage[style=ieee,
            backend=biber,
            giveninits=true,
            url=false,
            doi=false,
            eprint=false,
            isbn=false]{biblatex}
\AtEveryBibitem{\clearfield{pages}}

\title{\LARGE \bf
Gaussian Process-Augmented Model Predictive Contouring Control for 6-DoF Trajectory Tracking with a Holonomic AUV
}

\author{David Dörner$^{*1}$, Giancarlo Troni$^{2}$, Ignacio Torroba$^{1}$, Ivan Stenius$^{1}$%
\thanks{The authors are with $^{1}$ School of Engineering Sciences, KTH Royal Institute of Technology,
Stockholm, Sweden, {\tt\small \{ddorner,torroba,stenius\}@kth.se}, $^{2}$ The Monterey Bay Aquarium Research Institute (MBARI), Moss Landing, CA 95039, USA, {\tt\small gtroni@mbari.org}, $^{*}$ corresponding author.}%
\thanks{This work was partially supported by the Wallenberg AI, Autonomous
Systems and Software Program (WASP) funded by the Knut and Alice
Wallenberg Foundation and by the David and Lucille Packard Foundation.}
}%

\begin{document}

\maketitle
\thispagestyle{empty}
\pagestyle{empty}

\begin{abstract}
Small, holonomic \glspl{auv} enable close-range inspection of underwater structures from arbitrary viewpoints, requiring accurate position and attitude tracking under state and input constraints. \Gls{mpcc} naturally handles these requirements, but relies on an accurate dynamics model, which is difficult to obtain underwater.
In this paper, we present a nonlinear \gls{mpcc} in all six \glspl{dof} for a holonomic \gls{auv} and alleviate the inaccuracies in the dynamics by augmenting the nominal model employed with residual dynamics learned via \gls{gp} regression.
The learned residual is embedded in the optimal control problem through a zero-order robust optimization scheme, which keeps the controller real-time capable at \unit[25]{Hz}.
A simulation ablation shows that residual selection is essential when designing the controller, and simulation and in-water experiments with the \gls{mola} \gls{auv} validate the resulting controllers against a nominal \gls{mpcc} and a field-proven \gls{pid} controller.
\end{abstract}
\glsresetall

\section{INTRODUCTION}
Close-range inspection of underwater structures such as wrecks, pipelines or mines requires precise maneuvering and the ability to observe a target from multiple viewpoints.
Torpedo-shaped \glspl{auv}, such as the Kongsberg Hugin and the \gls{mbari} LRAUV, are well suited to surveying large areas, but their limited maneuverability makes these inspection tasks challenging for them.
\Glspl{rov} are commonly used for such tasks, but they require a support vessel and crew for operation.
Small, holonomic \glspl{auv}, such as the \gls{mola} \gls{auv} of \gls{mbari}, offer an autonomous alternative: they can control position and attitude independently, allowing them to maintain a desired viewing direction while moving around a target.
Exploiting this capability, however, requires a controller capable of following complex position and attitude references while respecting state and input constraints for safe operation near underwater structures.

\Gls{mpc}~\cite{rawlings2024Model} naturally incorporates these requirements by solving a constrained optimal control problem over a receding horizon.
However, its performance relies on an accurate model of the system dynamics, which is a challenge in itself for \glspl{auv}: the hydrodynamic parameters are hard to identify and change with environmental and mission-specific conditions, such as different payloads.
Recent works therefore apply learning-based methods to improve the model used in the \gls{mpc}~\cite{amer2025Empowering,bhat2026Online}, but they either restrict the dynamics to four \glspl{dof} or remain validated in simulation only.

\begin{figure}[t]
    \centering
    \includegraphics[width=0.9\linewidth]{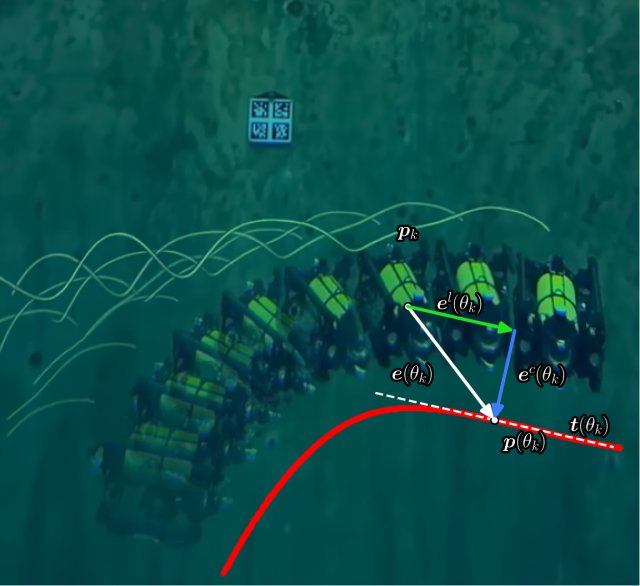}
    \caption{Composite of the \gls{mola} \gls{auv} during an experiment in the tank, overlaid with the error approximation of the MPCC over the prescribed trajectory (red): the position error $\vec{e}(\theta_k)$ between the vehicle position $\vec{p}_k$ and the path point $\vec{p}(\theta_k)$ is split into the approximated lag error $\vec{e}^l(\theta_k)$ along the tangent $\vec{t}(\theta_k)$ and the approximated contouring error $\vec{e}^c(\theta_k)$ in the normal plane.}
    \label{fig:mola_pic}
\end{figure}

In this paper, we close this gap with the following contributions:
\begin{itemize}
    \item We present a nonlinear \gls{mpcc} in all six \glspl{dof} that exploits the holonomic actuation of the \gls{auv} to track position and attitude references independently. To the best of our knowledge, this is the first real-time deployment of such a scheme on a holonomic \gls{auv}.
    \item We augment the nominal model with \gls{gp} residual dynamics, embed them into the \gls{mpcc} via a zero-order robust optimization scheme, and show in a simulation ablation that the choice of which residuals to learn is decisive for the resulting performance.
    \item We validate the controllers in-water on the \gls{mola} \gls{auv} against a well-tuned \gls{pid} controller and demonstrate improved attitude tracking on a lawnmower survey as well as a spiraling inspection maneuver.
\end{itemize}

\section{RELATED WORK}\label{sec:related_work}
\Gls{mpc} has attracted growing interest for \glspl{auv} in recent years~\cite{slawik2026Recent,wei2023Mpcbased}.
Often, the model is reduced to three \glspl{dof}, i.e., surge, sway, and yaw~\cite{tian2025Safe,shen2017Modified,shen2017Integrated}.
While reducing the model order speeds up the computation, depth, roll, and pitch control are equally important for inspection tasks.
Trajectory tracking for a fully actuated vehicle considering all states is proposed in~\cite{xu2024ThreeDimensional}, but, like the aforementioned works, it is verified in simulation only.
Experimental validation is provided in~\cite{cai2024Autonomous}, where \gls{mpc} controls surge, heave, and heading of an \gls{auv} with an attached manipulator, in~\cite{gao2024Model} for the position control of a holonomic \gls{auv}, and in~\cite{saback2020Nonlinear} for velocity control.
\Gls{mpc} was further used to dock a holonomic \gls{auv} to a moving docking station~\cite{vivekanandan2023Autonomous}, to move an \gls{rov} between waypoints~\cite{heshmati-alamdari2018Robusta}, also in a self-triggered fashion~\cite{heshmati-alamdari2014selftriggered}, and to mimic inspection tasks on a space station with a holonomic \gls{rov}~\cite{verhagen2026Validation}.
These works track waypoints or time-parameterized references.
\Gls{mpcc}~\cite{lam2010Model}, in contrast, follows a geometric path and optimizes the progress along it, which has enabled time-optimal driving~\cite{liniger2015Optimizationbased} and flight~\cite{romero2022Model,krinner2024MPCC}.
To the best of our knowledge, \gls{mpcc} has not been deployed on an underwater vehicle in all six \glspl{dof} yet.

Learning-based methods can be incorporated into \gls{mpc} in different ways~\cite{hewing2020LearningBased}: a) augmenting the dynamical model used in the optimization, b) learning the cost function, or c) using \gls{mpc} as a safety filter for a learned policy.
There is growing interest in incorporating the uncertainty of the learned model into the optimal control problem~\cite{scampicchio2025Gaussian}, e.g., to tighten chance constraints and increase the robustness of a race car~\cite{hewing2020Cautious}.
These methods are also increasingly applied in the underwater domain~\cite{macatangay2024Machine,hong2025Control}.
\Glspl{gp} combined with \gls{mpc} were used to predict the disturbances acting on an \gls{rov}~\cite{amer2025Empowering}; however, the righting moment of the \gls{rov} is leveraged to reduce the system dynamics to four \glspl{dof}.
In~\cite{bhat2026Online}, different methods for residual learning are compared, with the \gls{gp} residual performing best, but the work focuses on learning the hydrodynamic damping and validates the resulting \gls{ltv}-\gls{mpc} in simulation only.

In this work, we build upon~\cite{bhat2026Online} and use \glspl{gp} to learn a residual model, but without restricting it to the damping terms, and embed it into the optimal control problem following~\cite{lahr2026L4acados}.
Furthermore, we deploy a full nonlinear \gls{mpcc} in 3D~\cite{romero2022Model} and exploit the holonomic properties of the \gls{auv} to additionally track an attitude reference.
This allows us to orient the vehicle independently of its direction of travel, which is crucial during surveys and inspections.

\section{PRELIMINARIES}

\subsection{Problem Statement}
Consider the discrete-time nonlinear system
\begin{equation}
    \vec{x}(t+1) = \vec{f}(\vec{x}(t), \vec{u}(t)) + \vec{B}_d\,\vec{g}(\vec{x}(t),\vec{u}(t)) + \vec{w}(t),
    \label{eq:generic_system_equation}
\end{equation}
with state $\vec{x} \in \X \subseteq \R^{n_x}$ and input $\vec{u} \in \U \subseteq \R^{n_u}$.
The dynamics consist of a nominal part $\vec{f}$ derived from first principles, an unknown residual $\vec{g}: \R^{n_x}\times\R^{n_u} \rightarrow \R^{n_g}$ that captures unmodeled effects and parameter mismatch, and i.i.d.\ process noise $\vec{w}(t) \sim \N(\vec{0}, \vec{\Sigma}_w)$.
The matrix $\vec{B}_d \in \R^{n_x \times n_g}$ selects the states on which the residual acts.
We aim to control \cref{eq:generic_system_equation} optimally in a receding horizon fashion using \gls{mpc}~\cite{rawlings2024Model} as
\begin{equation}
    \begin{aligned}
        \pi(\vec{x}) &= \argmin_{\vec{u}} \sum_{k=0}^{N-1} \|\vec{e}_{\vec{x},k}\|_{\vec{Q}}^2 + \|\vec{u}_k\|_{\vec{R}}^2 + l_N(\vec{x}_N)\\
        \st\quad &\vec{x}_0 = \vec{x}(t)\\
        &\vec{x}_{k+1} = \vec{f}(\vec{x}_k, \vec{u}_k)\\
        &\vec{x}_k \in \X, \vec{u}_k \in \U,
    \end{aligned}
    \label{eq:standard_mpc}
\end{equation}
where $\vec{e}_{\vec{x}} = \vec{x}_{ref} - \vec{x}$ denotes the tracking error, $\vec{Q}$ and $\vec{R}$ are positive (semi-)definite weights on the tracking error and the control effort, and $l_N$ is the terminal cost.
Whereas the nominal part is known, we learn the residual from data.
From \cref{eq:generic_system_equation}, samples of the residual can be computed from consecutive state measurements as
\begin{equation}
    \vec{g}(\vec{x}(t),\vec{u}(t)) \approx \vec{B}_d^{\dagger}\left(\vec{x}(t+1) - \vec{f}(\vec{x}(t), \vec{u}(t))\right),
    \label{eq:res_computation}
\end{equation}
where $\vec{B}_d^{\dagger}$ is the Moore-Penrose pseudo-inverse.

\subsection{Notation}
We denote scalars as $a$, vectors as $\vec{a}$, and matrices as $\vec{A}$. $\|\vec{a}\|_{\vec{W}}^2 = \vec{a}\T\vec{W}\vec{a}$ denotes the weighted squared norm. $t$ denotes time, whereas $k$ denotes the prediction step within the \gls{mpc} horizon. For clarity, we omit the indices.

\subsection{AUV Dynamics}\label{sec:mola}
The nominal part of \cref{eq:generic_system_equation} follows from the well-known equations of motion of a marine craft~\cite{fossen2021Handbook},
\begin{equation}
    \begin{aligned}
        \dot{\vec{\eta}} &= \vec{J}(\vec{\eta})(\vec{\nu}_r + \vec{\nu}_c)\\
        \vec{M}\dot{\vec{\nu}}_r &+ \vec{C}(\vec{\nu}_r)\vec{\nu}_r + \vec{D}(\vec{\nu}_r)\vec{\nu}_r + \vec{g}(\vec{\eta}) + \vec{g}_0 = \vec{\tau},
    \end{aligned}
    \label{eq:general_eom}
\end{equation}
where $\vec{\eta}$ denotes the \gls{auv}'s pose in the world frame, $\vec{\nu}_r$ its velocity relative to the water expressed in the body frame, $\vec{\nu}_c$ the current velocity, and $\vec{J}(\vec{\eta})$ the Jacobian transforming body-frame velocities into the world frame.
The system is parameterized by the mass matrix $\vec{M} = \vec{M}_{RB} + \vec{M}_A$ and the Coriolis matrix $\vec{C} = \vec{C}_{RB} + \vec{C}_A$, both consisting of a rigid-body and an added-mass term, and by the hydrodynamic damping $\vec{D}$.
Gravity and buoyancy are expressed in $\vec{g}(\vec{\eta})$ and $\vec{g}_0$, and all external forces and moments, i.e., actuators and disturbances, in $\vec{\tau}$.
For the remainder of the paper, we use a body frame with \gls{frd} axes, i.e., the depth $z$ is positive downwards, and we represent the attitude by a unit quaternion $\vec{q}$ to avoid singularities, such that $\vec{\eta} = [x, y, z, \vec{q}\T]\T$ and $\vec{\nu} = [u, v, w, p, q, r]\T$.
Identifying every parameter in these matrices is challenging due to the inherent nonlinearities and cross-couplings~\cite{bhat2026Online}.
We simplify the model by assuming a symmetric vehicle body and a diagonal damping matrix $\vec{D}$, and we assume no known current, hence $\vec{\nu}_c = \vec{0}$ and $\vec{\nu}_r = \vec{\nu}$.
Solving the second line of \cref{eq:general_eom} for the acceleration yields
\begin{equation}
    \dot{\vec{\nu}}  = \vec{M}^{-1}\left(\vec{\tau} - \vec{C}(\vec{\nu})\vec{\nu} - \vec{D}(\vec{\nu})\vec{\nu} - \vec{g}(\vec{\eta}) - \vec{g}_0\right),
    \label{eq:eom_acceleration}
\end{equation}
which, together with the kinematics in \cref{eq:general_eom} and a discretization with the sampling time $\Delta t$, constitutes $\vec{f}$ in \cref{eq:generic_system_equation} with $\vec{x} = [\vec{\eta}\T, \vec{\nu}\T]\T$ and $\vec{u} = \vec{\tau}$.

\section{MODEL PREDICTIVE CONTOURING CONTROL}\label{sec:mpcc}

\subsection{Contouring and Lag Error}
Instead of tracking a time-parameterized reference as in \cref{eq:standard_mpc}, \gls{mpcc} follows a geometric reference path $\vec{p}(\theta)$ parameterized by its arc length $\theta$ and treats the progress along the path as an additional decision variable.
The distance between the current position $\vec{p}_k$ and the path is the distance to the closest point on the path, whose computation is itself an optimization problem.
Following~\cite{romero2022Model}, we approximate it using the path parameter $\theta_k$ at prediction step $k$, as illustrated in \cref{fig:mola_pic}.
Let $\vec{t}(\theta_k)$ be the unit tangent of the path at $\vec{p}(\theta_k)$ and $\vec{e}(\theta_k) = \vec{p}(\theta_k) - \vec{p}_k$ the position error.
Then, the lag error is the projection of the position error onto the tangent,
\begin{equation}
    \vec{e}^l(\theta_k) = \left(\vec{t}(\theta_k)\T\vec{e}(\theta_k)\right)\vec{t}(\theta_k) = e^l(\theta_k)\,\vec{t}(\theta_k),
    \label{eq:lag_error}
\end{equation}
and the contouring error is the remaining component in the normal plane of the path,
\begin{equation}
    \vec{e}^c(\theta_k) = \vec{e}(\theta_k) - e^l(\theta_k)\,\vec{t}(\theta_k).
    \label{eq:contouring_error}
\end{equation}
Their weighted norms follow as $\|\vec{e}^l\|_{Q_l}^2 = Q_l\,(e^l)^2$ and $\|\vec{e}^c\|_{\vec{Q}_c}^2 = \vec{e}^{c\mathrm{T}}\vec{Q}_c\,\vec{e}^c$.


\subsection{Optimal Control Problem}
The complete \gls{mpcc} problem follows as
\begin{equation}
    \begin{aligned}[t]
        \pi(\vec{x}) &= \argmin_{\vec{u},\Delta v_{\theta}} \sum_{k=0}^{N-1} \|\vec{e}^c_k\|_{\vec{Q}_c}^2
        + \|\vec{e}^l_k\|_{Q_l}^2 \\
        &\qquad + \|e_{v_{\theta},k}\|_{Q_v}^2
        + \|\vec{e}_{\vec{q},k}\|_{\vec{Q}_q}^2
        + \|\Delta\vec{q}_k\|_{\vec{Q}_{\Delta\vec{q}}}^2 \\
        &\qquad + \|\Delta\vec{u}_k\|_{\vec{R}}^2 + l_N(\vec{x}_N)\\
    \st\quad &\vec{x}_0 = \vec{x}(t),\ \theta_0 = \theta(t),\ v_{\theta,0} = v_{\theta}(t)\\
        &\vec{x}_{k+1} = \vec{f}(\vec{x}_k, \vec{u}_k)\\
        &\theta_{k+1} = \theta_k + v_{\theta,k}\Delta t\\
        &v_{\theta,k+1} = v_{\theta,k} + \Delta v_{\theta,k}\Delta t\\
        &\vec{x}_k \in \X,\quad \underline{\vec{u}} \leq \vec{u}_k \leq \widebar{\vec{u}}\\
        &0 \leq v_{\theta,k} \leq \widebar{v}_{\theta},\quad
        \underline{\Delta v}_{\theta} \leq \Delta v_{\theta,k} \leq \widebar{\Delta v}_{\theta}\\
        &h(\vec{x}_k) \leq 0,
    \end{aligned}
    \label{eq:mpcc}
\end{equation}
with the terminal cost $l_N = \|\vec{e}^c_N\|_{\vec{Q}_c}^2 + \|\vec{e}^l_N\|_{Q_l}^2 + \|\vec{e}_{\vec{q},N}\|_{\vec{Q}_q}^2$ for stability~\cite{chen1998Quasi-infinite}.
The progress along the path is encouraged by the term $e_{v_{\theta}} = v_{\theta,ref} - v_{\theta}$: choosing $v_{\theta,ref} \gg \widebar{v}_{\theta}$ recovers the linear progress reward of~\cite{romero2022Model} while keeping the cost convex, which improves the solver performance.
Leveraging the holonomic actuation of the \gls{auv}, we add the attitude error $\vec{e}_{\vec{q}} = \vec{q}_{ref} \otimes \vec{q}^{-1}$, where $\otimes$ denotes the quaternion product and $\vec{q}^{-1}$ the conjugate of the current attitude, and penalize its deviation from the identity quaternion.
For numerical stability, we additionally penalize the change in attitude between consecutive steps, $\Delta\vec{q}_k$, which acts as a soft angular rate limit~\cite{romero2022Model}, as well as the change in control input $\Delta\vec{u}_k = \vec{u}_k - \vec{u}_{k-1}$ to avoid jerky motions.
To prevent noisy control inputs, the progress speed $v_{\theta}$ is treated as a virtual state and its rate of change $\Delta v_{\theta}$ as the optimized control variable.
Finally, the nonlinear state constraint $h(\vec{x}) = \|\vec{e}_c\| - r \leq 0$ constraints the deviation from the trajectory's center line with radius $r$.

\subsection{Reference Trajectory}
The reference for the \gls{auv} is generated from position and orientation waypoints.
The positions are connected by cubic splines, which yield the path $\vec{p}(\theta)$ and its arc length $\theta$.
The attitude reference $\vec{q}_{ref}(\theta)$ is obtained by spherical linear interpolation (SLERP)~\cite{shoemake1985Animating} between the waypoint orientations, parameterized by the same arc length such that position and attitude references are aligned along the path.

\section{\gls{gp}-\gls{mpcc}}\label{sec:gp}
There are different ways to learn the residual in \cref{eq:res_computation}~\cite{bhat2026Online}.
In this work, we use \glspl{gp}~\cite{rasmussen2008Gaussian}, which provide a measure of uncertainty alongside the mean prediction~\cite{hewing2020Cautious}.
Using \cref{eq:res_computation}, the training targets are computed from recorded state and input data as
\begin{equation}
    \vec{y}(t) = \vec{B}_d^{\dagger}\left(\vec{x}(t+1) - \vec{f}(\vec{x}(t), \vec{u}(t))\right),
    \label{eq:data_computation}
\end{equation}
with the regression inputs $\vec{z}(t) = \left[\vec{x}(t)\T, \vec{u}(t)\T\right]\T$.

\subsection{Gaussian Processes}
Given the training inputs $\vec{z}$ and targets $\vec{y}$, the posterior mean and variance of a \gls{gp} at a query point $\vec{z}_*$ are
\begin{equation}
    \begin{aligned}
        \vec{\mu}^g(\vec{z}_*) &= \vec{K}_{\vec{z}_*,\vec{z}}\left(\vec{K}_{\vec{z},\vec{z}} + \sigma_n^2 \vec{I}\right)^{-1} \vec{y}\\
        \vec{\Sigma}^g(\vec{z}_*) &= \vec{K}_{\vec{z}_*, \vec{z}_*} - \vec{K}_{\vec{z}_*,\vec{z}}\left(\vec{K}_{\vec{z},\vec{z}} + \sigma_n^2 \vec{I}\right)^{-1} \vec{K}_{\vec{z},\vec{z}_*},
    \end{aligned}
    \label{eq:gp_def}
\end{equation}
where $\vec{K}_{\cdot,\cdot}$ denotes the kernel matrix evaluated between the respective inputs and $\sigma_n^2$ the noise variance.
The kernel encodes prior assumptions on the residual, e.g., its smoothness~\cite{rasmussen2008Gaussian}, and its hyperparameters are optimized by maximizing the marginal likelihood of the training data.

\subsection{GP-MPCC Formulation}\label{ssec:gp_mpc}
\begin{figure}
    \centering
    \includegraphics[width=0.95\linewidth]{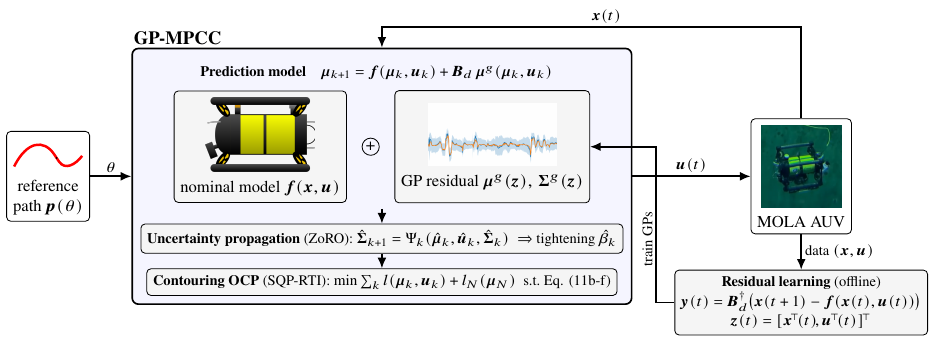}
    \caption{Overview of the proposed \gls{gp}-\gls{mpcc} framework: an offline-learned \gls{gp} residual augments the nominal \gls{auv} model, with propagated uncertainty tightening the constraints of an online contouring MPC}
    \label{fig:gp-mpcc_framework}
\end{figure}

Following~\cite{lahr2026L4acados}, we embed the learned residual into the \gls{mpcc} formulation \cref{eq:mpcc} by replacing the nominal dynamics with the mean dynamics and propagating the resulting uncertainty along the horizon:
\begin{subequations}
    \begin{align}
        \min_{\vec{u},\vec{\mu},\vec{\Sigma}}\quad &\sum_{k=0}^{N-1}l(\vec{\mu}_k, \vec{u}_k) + l_N(\vec{\mu}_N) \\
        \st\quad &\vec{\mu}_0 = \vec{x}(t),\quad \vec{\Sigma}_0 = \vec{0},\label{eq:sigma_0_0_ass}\\
        &\vec{\mu}_{k+1} = \vec{f}(\vec{\mu}_k, \vec{u}_k) + \vec{B}_d\,\vec{\mu}^g(\vec{\mu}_k, \vec{u}_k),\label{eq:gp_mean_dyn}\\
        &\vec{\Sigma}_{k+1} = \Psi_k(\vec{\mu}_k, \vec{u}_k, \vec{\Sigma}_k),\label{eq:cov_dyn_constraint}\\
        &h(\vec{\mu}_k) + \beta(\vec{\mu}_k, \vec{\Sigma}_k) \leq 0,\label{eq:tightened_constraint}\\
        &h(\vec{\mu}_N) + \beta(\vec{\mu}_N, \vec{\Sigma}_N) \leq 0,\label{eq:tightened_constraint}
    \end{align}
    \label{eq:gp_mpc}
\end{subequations}
where $l$ collects the stage cost terms of \cref{eq:mpcc}, the remaining decision variables and constraints of \cref{eq:mpcc} are omitted for brevity, and $\vec{\mu}^g$ and $\vec{\Sigma}^g$ are the \gls{gp} mean and variance from \cref{eq:gp_def} evaluated at $\vec{z} = [\vec{\mu}_k\T, \vec{u}_k\T]\T$.
The state covariance is propagated by linearizing the dynamics,
\begin{equation}
    \Psi_k(\vec{\mu}_k, \vec{u}_k, \vec{\Sigma}_k) = \vec{A}_k\vec{\Sigma}_k\vec{A}_k\T + \vec{B}_d\vec{\Sigma}^g(\vec{\mu}_k,\vec{u}_k)\vec{B}_d\T + \vec{\Sigma}_w,
    \label{eq:cov_propagation}
\end{equation}
with the Jacobian of the nominal and the \gls{gp} dynamics
\begin{equation}
    \vec{A}_k = \left.\frac{\partial \vec{f}(\vec{x},\vec{u}_k)}{\partial \vec{x}} + \vec{B}_d\frac{\partial \vec{\mu}^g(\vec{x},\vec{u}_k)}{\partial \vec{x}}\right|_{\vec{x}=\vec{\mu}_k}.
    \label{eq:dynamics_jacobian}
\end{equation}
The back-off term $\beta$ tightens the constraints according to the predicted uncertainty, e.g., $\beta_i = \gamma\sqrt{\nabla h_i\T\vec{\Sigma}_k\nabla h_i}$ for each constraint $h_i$, where $\gamma$ sets the desired confidence level~\cite{hewing2020Cautious}.
Note that \cref{eq:gp_mpc} has the covariances $\vec{\Sigma}_k$ as decision variables, which is computationally expensive~\cite{lahr2026L4acados}.
Following~\cite{lahr2023Zeroorder}, we instead propagate the covariance outside of the \gls{sqp} iterations based on the previous solution $(\hat{\vec{\mu}}_k, \hat{\vec{u}}_k)$ in a \gls{zoro} fashion,
\begin{equation}
    \hat{\vec{\Sigma}}_{k+1} = \Psi_k(\hat{\vec{\mu}}_k, \hat{\vec{u}}_k, \hat{\vec{\Sigma}}_k),
    \label{eq:cov_prop_approx}
\end{equation}
and evaluate the tightening $\hat{\beta}_k = \beta(\hat{\vec{\mu}}_k,  \hat{\vec{u}}_k, \hat{\vec{\Sigma}}_k)$ and $\hat{\beta}_N = \beta(\hat{\vec{\mu}}_N, \hat{\vec{\Sigma}}_N)$ with the propagated covariances.
The optimal control problem then reduces to \cref{eq:gp_mpc} without the covariance dynamics \cref{eq:cov_dyn_constraint} and with the fixed tightening $\hat{\beta}_k$ in \cref{eq:tightened_constraint}.
It has the same structure as the nominal problem \cref{eq:mpcc} and is solved with the same \gls{sqp} machinery, where the \gls{gp} mean and its Jacobian are evaluated externally and passed to the solver~\cite{lahr2026L4acados}. \Cref{fig:gp-mpcc_framework} summarizes the proposed framework.

\section{EXPERIMENTS}\label{sec:validation}
\begin{figure}[h]
    \centering
    \includegraphics[width=0.95\linewidth]{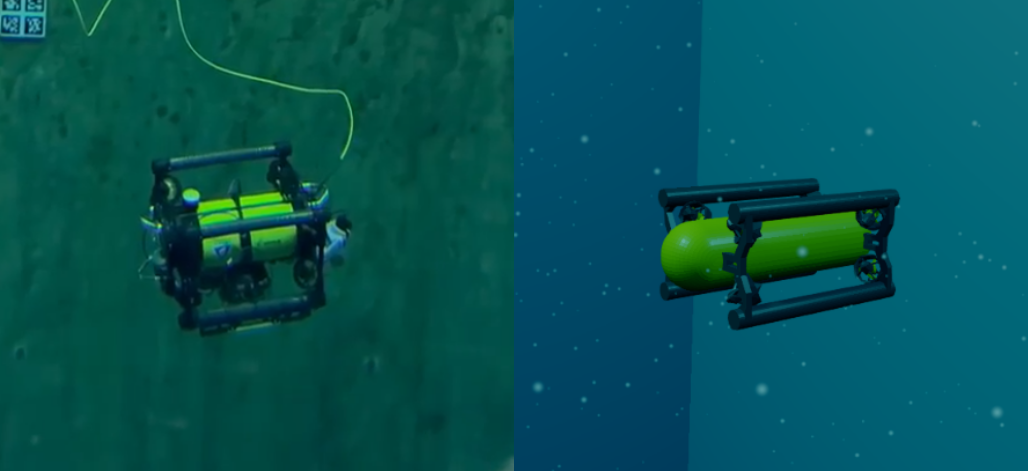}
    \caption{The \gls{mola} \gls{auv} and its simulation in Stonefish.}
    \label{fig:sim_and_real}
\end{figure}

We validate the proposed method with the \gls{mola} \gls{auv} first in simulation and then in hardware experiments (\cref{fig:sim_and_real}).
In both cases, the setup is kept the same, allowing for a seamless transition from simulation to the real vehicle: the controller, the communication, and the residual learning pipeline are identical, and only the vehicle interface differs.

Since the hull of \gls{mola} \gls{auv} resembles a BlueROV2 and no identified hydrodynamic model of the vehicle is available, we parameterize the nominal model $\vec{f}$ with the known weight and the BlueROV2 coefficients of~\cite{vonbenzon2022OpenSource} as used in~\cite{verhagen2026Validation}.
The resulting mismatch between the nominal model and the true vehicle is precisely what the learned residual is meant to compensate.
The controller is implemented in Python using \texttt{acados}~\cite{verschueren2022acados} for the nominal \gls{mpcc} and \texttt{L4acados}~\cite{lahr2026L4acados} for the \gls{zoro} implementation of the \gls{gp}-\gls{mpcc} presented in \cref{ssec:gp_mpc}.
To guarantee real-time feasibility, the optimal control problem is solved with the \gls{rti} scheme~\cite{diehl2005Realtime}, i.e., one \gls{sqp} iteration per sampling instant, warm-started from the previous solution.
The controller runs at \unit[25]{Hz} with a prediction horizon of $N=40$ steps on an Ubuntu 24.04 notebook with an Intel i7 CPU and \unit[64]{GB} of RAM.
Communication with the vehicle, both in simulation and in the experiments, runs via \gls{lcm}~\cite{huang2010LCM}.
The \gls{mpcc} weights were tuned empirically and kept the same for the simulation and the experimental validation.

\subsection{Residual Learning}
The residuals of the six body velocities in $\vec{\nu}$ are modeled as independent \glspl{gp}, each with a \gls{se} kernel with \gls{ard}, implemented in \texttt{GPyTorch}~\cite{gardner2018GPyTorch}; $\vec{B}_d$ selects the velocity states accordingly, and the regression inputs are the body velocities and the control inputs.
Training data is collected in open loop by exciting each axis first individually with increasing effort and subsequently in a coupled manner.
The recorded data is sub-sampled to 100 uniformly spread data points per \gls{gp}, which keeps the inference cost within the real-time budget.
Separate models are trained on the same maneuvers for the simulation and for the real vehicle, respectively.

\begin{figure}[t]
    \centering
    \includegraphics[width=0.95\linewidth]{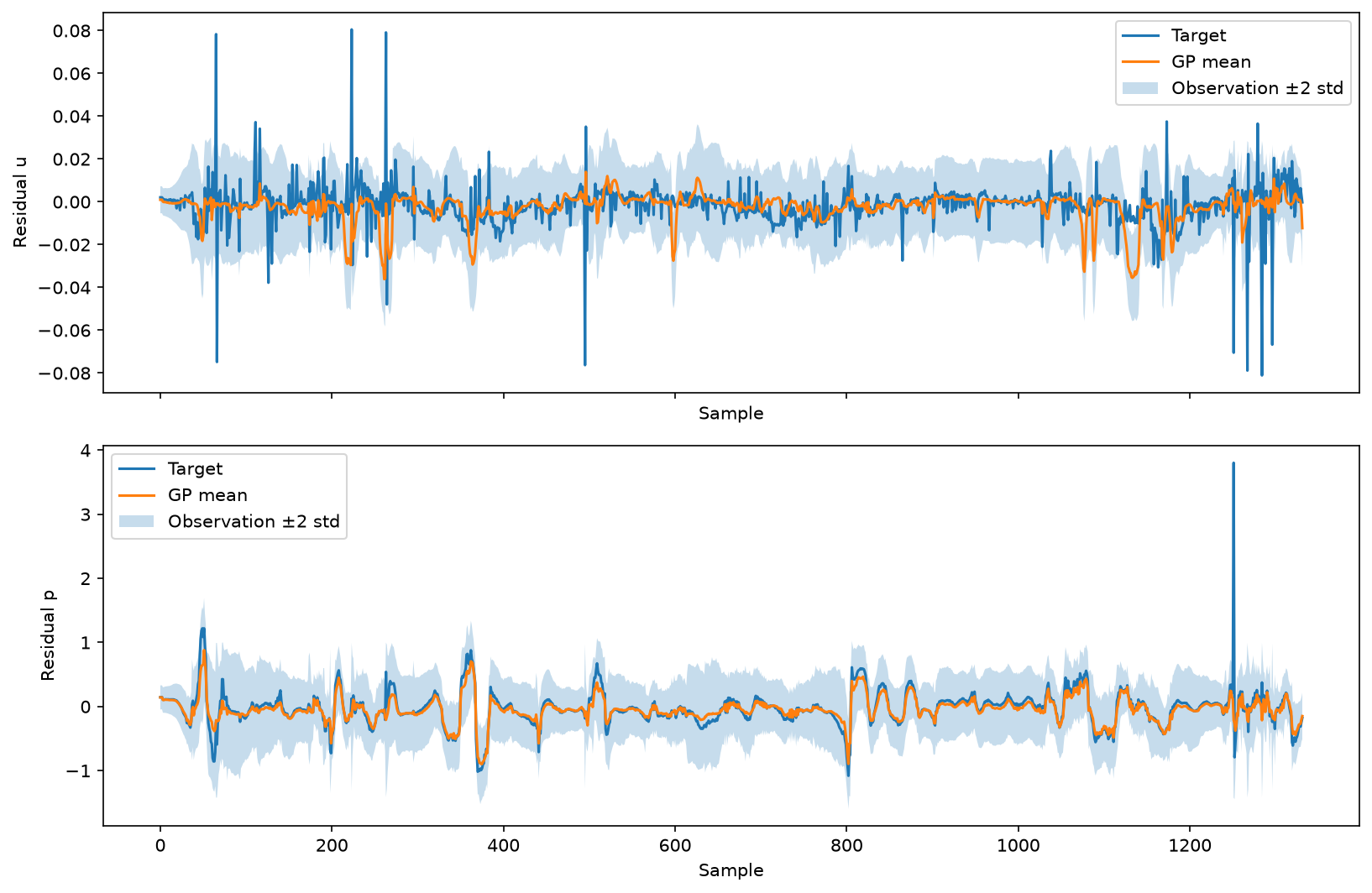}
    \caption{\Gls{gp} prediction (mean and $\pm 2\sigma$ band) of the surge residual $u$ (top) and the roll rate residual $p$ (bottom) against the measured residual (target) during an in-water experiment. The linear residual is at noise level, whereas the angular residual exhibits structure that the \gls{gp} captures.}
    \label{fig:test_predictions}
\end{figure}

\begin{figure*}[t]
    \centering
    \includegraphics[width=1\linewidth]{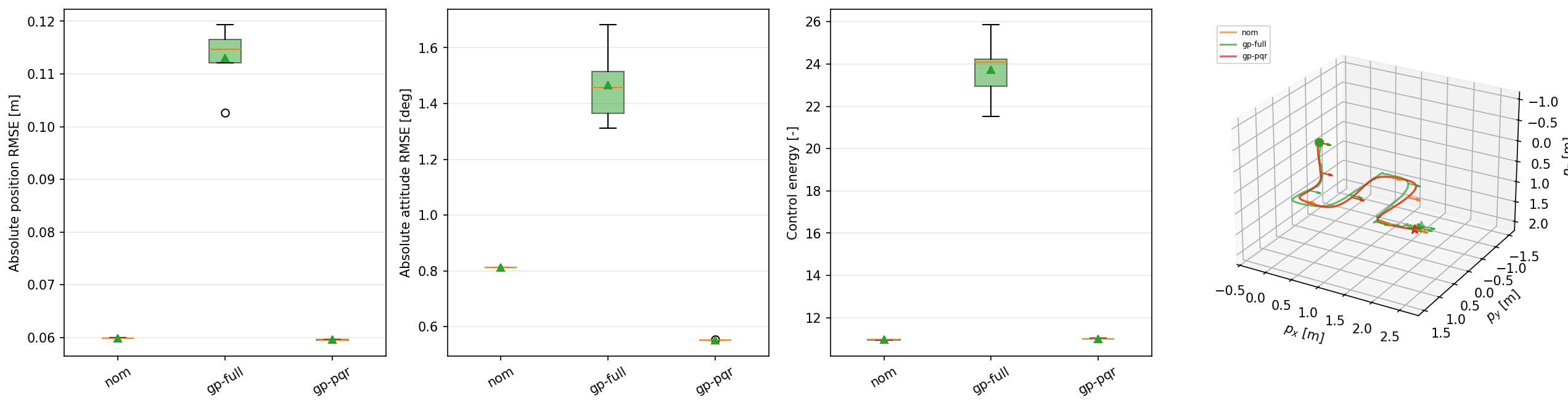}
    \caption{Ablation study with a simulated lawnmower pattern, five runs per controller: position and attitude \acrshort{rmse}, control energy, and the trajectories of one representative run. The GP-augmented controller is split into linear and angular components to showcase the deterioration in performance caused by the linear GPs.}
    \label{fig:box_plots_Stonefish}
\end{figure*}

\Cref{fig:test_predictions} shows the \gls{gp} predictions of a representative linear and angular residual during one of the in-water experiments.
The linear residual is dominated by measurement noise and offers little structure to exploit, whereas the angular residual exhibits a clear structure that the \gls{gp} captures well.
The consequence of this is clearly shown in the ablation experiment in \Cref{fig:box_plots_Stonefish}, where the linear residuals predictions mainly inject noise into the model, degrading the controller performance for the gp-full. 
We therefore focus on the angular residuals in the remaining experiments.

\begin{figure*}[t]
    \centering
    \includegraphics[width=1\linewidth]{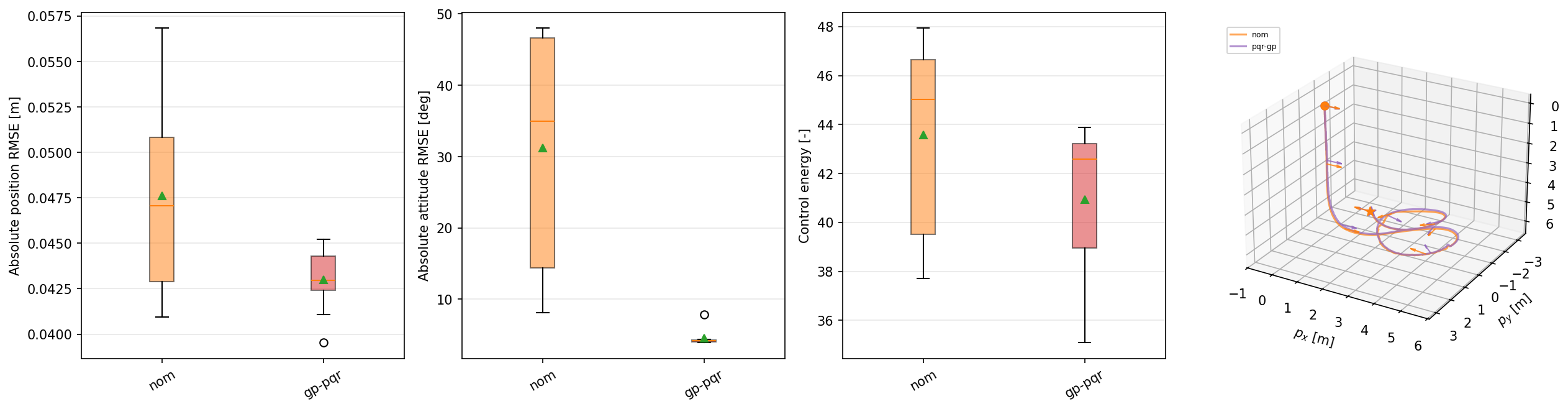}
    \caption{Simulated inspection spiral, ten runs per controller: position and attitude \acrshort{rmse}, control energy, and the trajectories of one representative run.}
    \label{fig:box_plots_Stonefish_spiral}
\end{figure*}
\subsection{Evaluation Metrics}
Since the reference is a geometric path, at every step $i$ we evaluate the tracking error against the closest point on the position spline rather than against a time-indexed reference.
For each recorded state $\vec{x}_i = [\vec{\eta}_i\T, \vec{\nu}_i\T]\T$ with pose $\vec{\eta}_i = [\vec{p}_i\T, \vec{q}_i\T]\T$, we determine the closest point on the position spline,
\begin{equation}
    \theta_i^* = \argmin_{\theta} \|\vec{p}(\theta) - \vec{p}_i\|_2,
    \label{eq:closest_point}
\end{equation}
and evaluate the reference pose $[\vec{p}(\theta_i^*)\T, \vec{q}_{ref}(\theta_i^*)\T]\T$ at that arc length.
The position and attitude errors then follow, consistent with \cref{eq:mpcc}, as
\begin{equation}
    \vec{e}_{\vec{p},i} = \vec{p}_i - \vec{p}(\theta_i^*), \qquad
    \vec{e}_{\vec{q},i} = \vec{q}_{ref}(\theta_i^*) \otimes \vec{q}_i^{-1},
    \label{eq:pose_error}
\end{equation}
where we use the Euclidean norm $\|\vec{e}_{\vec{p},i}\|_2$ and the rotation angle $\varphi_i = 2\arccos|e_{\vec{q},i,w}|$ of the error quaternion, with $e_{\vec{q},i,w}$ its scalar part.
Both are aggregated over a run as the \gls{rmse}.
Lastly, we report the control energy as the sum of the absolute commanded control effort, $\sum_i \|\vec{\tau}\|_i$, as a proxy for the energy consumption.

\begin{figure*}[t]
    \centering
    \includegraphics[width=1.0\linewidth]{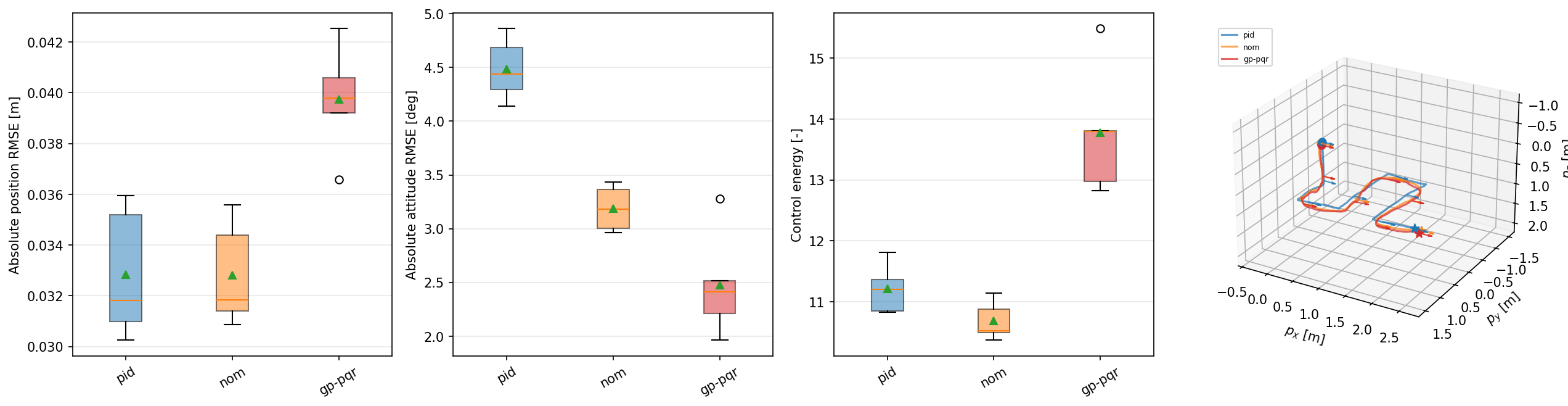}
    \caption{In-water lawnmower pattern, five runs per controller: position and attitude \acrshort{rmse}, control energy, and the trajectories of one representative run. Boxes span the quartiles and triangles mark the mean.}
    \label{fig:box_plots_tank_lawnmower}
\end{figure*}
\subsection{Simulation Study}
The controllers are first analyzed in the Stonefish simulator~\cite{cieslak2019Stonefish} in two scenarios.
The first is a constant heading lawnmower pattern, similar to a common survey in the field, on which we compare the nominal \gls{mpcc} (nom) against the angular-residual (gp-pqr) and the full-state (gp-full) \gls{gp}-\gls{mpcc} over five runs each.
The second is an upward spiral around a virtual object of interest, during which the vehicle keeps pointing towards the center while gradually increasing its pitch angle, mimicking a close-range inspection.
This maneuver demands continuous attitude changes and is repeated ten times with the nominal and the angular-residual \gls{mpcc}.

\subsection{Hardware Validation}
We validate the presented methods on the \gls{mola} \gls{auv} shown in \cref{fig:mola_pic}.
The \gls{mola} \gls{auv} is equipped with an \gls{imu}, a pressure sensor, and a \gls{dvl}, which are fused by an \gls{ekf} running on an onboard NVIDIA Orin NX for state estimation.
The \gls{mpcc} runs topside on the same notebook as in the simulation study and sends the commanded normalized body-frame efforts $\vec{\tau}$ to the vehicle, where the onboard thrust allocation distributes them to the individual thrusters.
The experiments were conducted in a $13 \times 9 \times \unit[10]{m}$ saltwater test tank.
We compare the nominal \gls{mpcc} as well as the angular-residual \gls{gp}-\gls{mpcc} against the well-tuned \gls{pid} controller running on the \gls{mola} \gls{auv}, which has proven itself during multiple field deployments and tracks the same waypoints sequentially as pose setpoints.
Each controller performs the lawnmower pattern five times.
As a final test of the method's deployability in a real scenario, the controller execute the more challenging inspection spiral.

\begin{figure}[t]
    \centering
    \includegraphics[width=0.85\linewidth]{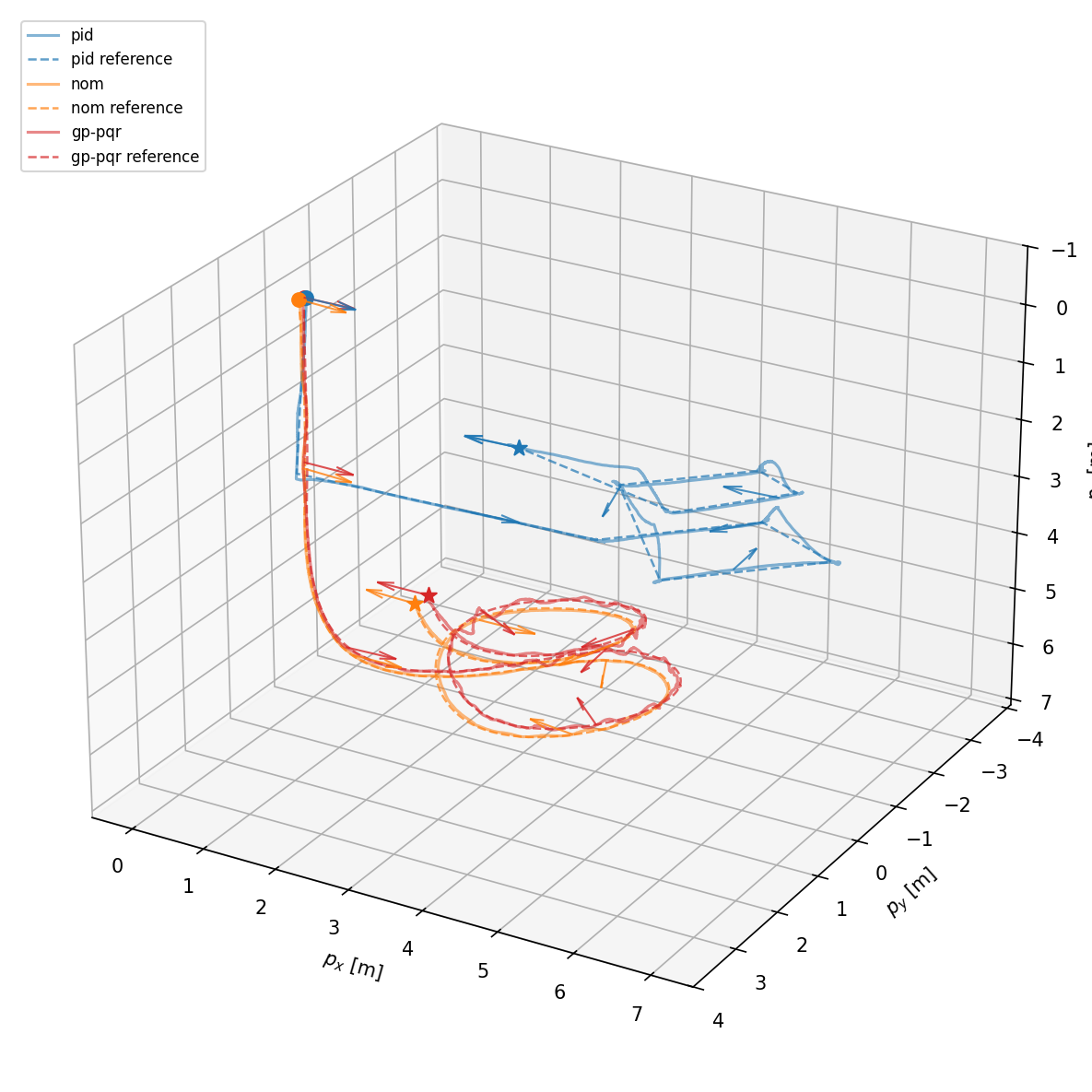}
    \caption{Upward inspection spiral in the test tank: the \gls{mpcc} variants follow the spline reference, whereas the \gls{pid} controller moves between the waypoints one \gls{dof} at a time.}
    \label{fig:tank_spiral}
\end{figure}
\section{RESULTS}\label{sec:discussion}

\subsection{Simulation}
\Cref{fig:box_plots_Stonefish} summarizes the lawnmower runs in simulation.
The angular-residual \gls{gp}-\gls{mpcc} reduces the attitude \gls{rmse} from $0.81^\circ$ to $0.55^\circ$ (32\%) at an unchanged position \gls{rmse} of \unit[0.06]{m} and unchanged control energy.
The full-state variant, on the other hand, roughly doubles the position and attitude errors as well as the control energy, and its trajectory visibly oscillates around the reference, confirming that the noisy linear residuals deteriorate the closed-loop performance.
The inspection spiral in \cref{fig:box_plots_Stonefish_spiral} shows the benefit of the learned angular residual much more clearly.
The mismatch in the rotational dynamics, which had a small impact on the lawnmower pattern with constant attitude, becomes dominant when the reference demands continuous attitude changes.
With the nominal model, the attitude \gls{rmse} ranges from $8^\circ$ to $48^\circ$ with a mean of $31^\circ$, i.e., the controller does not reliably hold the commanded attitude while circling the object.
The angular-residual \gls{gp}-\gls{mpcc}, in contrast, tracks the attitude consistently with an \gls{rmse} of $4.5^\circ$, while the position \gls{rmse} decreases from \unit[0.048]{m} to \unit[0.043]{m} and the control energy by 6\%.

\subsection{In-Water Experiments}
\Cref{fig:box_plots_tank_lawnmower} shows the aggregated metrics of the lawnmower experiments with the \gls{mola} \gls{auv}.
The nominal \gls{mpcc} matches the position \gls{rmse} of the \gls{pid} controller of \unit[0.033]{m}, while reducing the attitude \gls{rmse} from $4.5^\circ$ to $3.2^\circ$ at a slightly lower control energy.
Adding the angular residual further reduces the attitude \gls{rmse} to $2.4^\circ$, i.e., by 47\% compared to the \gls{pid} and by 25\% compared to the nominal \gls{mpcc}, at the cost of a higher position \gls{rmse} of \unit[0.040]{m} and 28\% more control energy than the nominal \gls{mpcc}.
The higher control energy of the \gls{gp}-augmented model stems from more aggressive control actions; since the \gls{mpcc} weights are kept the same across all experiments, a higher weight on the input rate can mitigate this effect.

\Cref{fig:tank_spiral} shows the inspection spiral in the test tank.
Both the nominal and the angular-residual \gls{gp}-\gls{mpcc} follow the spiral closely, whereas the \gls{pid} controller, which regulates each \gls{dof} independently towards the next waypoint, resorts to a piecewise motion that does not resemble the intended inspection path.
Since the \gls{pid} controller does not follow the spline reference, a quantitative comparison is not meaningful for this maneuver.

\subsection{Discussion}
The results in simulation and in the test tank show consistent qualitative behavior.
Compensating the angular residuals improves the attitude tracking, whereas using a \gls{gp} for every \gls{dof} degrades the performance in our setup, so that domain knowledge about where the nominal model is deficient is crucial to deploy learning-based methods sensibly.
The benefit of the learned residual grows with the complexity of the maneuver: it is moderate on the lawnmower pattern and substantial on the inspection spiral.
Furthermore, the nominal and the angular-residual \gls{mpcc} perform similar to or better than the \gls{pid} controller on the lawnmower pattern at a comparable control energy, while the \gls{pid} fails to follow the more complex spiraling inspection motion.
It is worth noting that considerably less engineering time was spent on integrating and tuning the \gls{mpcc}-based controllers than on the \gls{pid} controller, which has matured over multiple field deployments.
Beyond the tracking performance, the \gls{mpcc} formulation directly incorporates state and actuator constraints, which the \gls{pid} controller cannot.

\section{CONCLUSIONS}\label{sec:conclusion}
In this paper, we presented a 6-\gls{dof} \gls{mpcc} scheme for a holonomic \gls{auv} that tracks position and attitude references independently, and augmented its model with \gls{gp}-learned residual dynamics.
Simulation and in-water experiments showed that a sensible selection of the learned residuals improves the tracking performance, whereas learning all residuals is detrimental.
Compensating the angular residuals halves the attitude error of a field-proven \gls{pid} controller on a lawnmower survey and, on a spiraling inspection maneuver in simulation, reduces the attitude error of the nominal \gls{mpcc} by 85\% at a lower control energy.
The \gls{mpcc} approaches match or outperform the \gls{pid} controller on the lawnmower pattern and follow an inspection maneuver that the \gls{pid} cannot execute, while requiring less engineering effort.

Several avenues open up from here.
While training the \glspl{gp} is lightweight, online learning becomes crucial during longer deployments to reflect changes in the dynamics over time, and active learning schemes could improve the coverage of the state space.
Navigation is an essential factor during field deployments, and state estimation underwater is inherently uncertain. Incorporating this uncertainty into the propagation scheme of the \gls{gp}-\gls{mpcc} is a natural extension.
Improving the constraint tightening would further increase the robustness and allow for maneuvers where constraint satisfaction is critical, i.e., the close inspection of fragile underwater structures.
Finally, since the \gls{mpcc} readily incorporates constraints and geometric references, it enables further capabilities such as obstacle avoidance or seafloor 


\section*{ACKNOWLEDGMENT}
In accordance with the IEEE policy on AI-generated content, the authors disclose that Claude (Anthropic) was used for coding assistance during the implementation and for language editing of the manuscript.

\printbibliography

\end{document}